\documentclass[conference]{IEEEtran}
\IEEEoverridecommandlockouts
\usepackage{cite}
\usepackage{amsmath,amssymb,amsfonts}
\usepackage{algorithm}
\usepackage{algorithmic}
\usepackage{graphicx}
\usepackage{textcomp}
\usepackage{xcolor}

\usepackage{siunitx}
\usepackage{multirow}
\usepackage{setspace}
\usepackage{url}

\def\BibTeX{{\rm B\kern-.05em{\sc i\kern-.025em b}\kern-.08em
    T\kern-.1667em\lower.7ex\hbox{E}\kern-.125emX}}
\begin{document}

\title{
Joint Layout Analysis, Character Detection and Recognition for Historical Document Digitization
%\\
}

\author{\IEEEauthorblockN{Weihong Ma, Hesuo Zhang, Lianwen Jin, Sihang Wu, Jiapeng Wang}
\IEEEauthorblockA{
South China University of Technology\\
\{scutmaweihong, hesuozhang, lianwen.jin, hangscut, scutjpwang\} @gmail.com}
\and
\IEEEauthorblockN{Yongpan Wang}
\IEEEauthorblockA{
Alibaba Group \\
yongpan@alibaba-inc.com}
}
\renewcommand{\algorithmicrequire}{\textbf{Input:}} 
\renewcommand{\algorithmicensure}{\textbf{Output:}}
\maketitle

\begin{abstract}
In this paper, we propose an end-to-end trainable framework for restoring historical documents content that follows the correct reading order.
In this framework, two branches named character branch and layout branch are added behind the feature extraction network.
The character branch localizes individual characters in a document image and recognizes them simultaneously.
Then we adopt a post-processing method to group them into text lines.
The layout branch based on fully convolutional network outputs a binary mask.
We then use Hough transform for line detection on the binary mask and combine character results with the layout information to restore document content.
These two branches can be trained in parallel and are easy to train.
Furthermore, we propose a re-score mechanism to minimize recognition error.
Experiment results on the extended Chinese historical document MTHv2 dataset demonstrate the effectiveness of the proposed framework.
\end{abstract}


\section{Introduction}

In the course of thousands of years, a large number of historical documents containing valuable information on historical knowledge and literary arts are left behind.
After years of storage, historical document collection has encountered serious degradation, including staining, tearing and ink seepage.
The problem of how to preserve historical documents has attracted the attention of many researchers in recent years.
One efficient way is to use a document digitization system\cite{ding2004document}, which can protect printed paper documents from the effect of direct manipulation for the purposes of consultation, exchange and remote access\cite{xie2019weakly}.

In general, a document digitization system consists of two principal stages: layout analysis and text recognition.
Document layout analysis separates a document image into regions of interest, which is the prerequisite step for text recognition.
The main challenge in historical documents is the complex background and various layout situations.
Document text recognition methods can be divided into character-based and sequence-based.
Character-based recognition methods mostly first localize individual characters and then recognize and group them into text line.
Sequence-based methods regress text line and solve text recognition as a sequence labeling problem.

\begin{figure}
    \centering
    \includegraphics[width=0.4\textwidth]{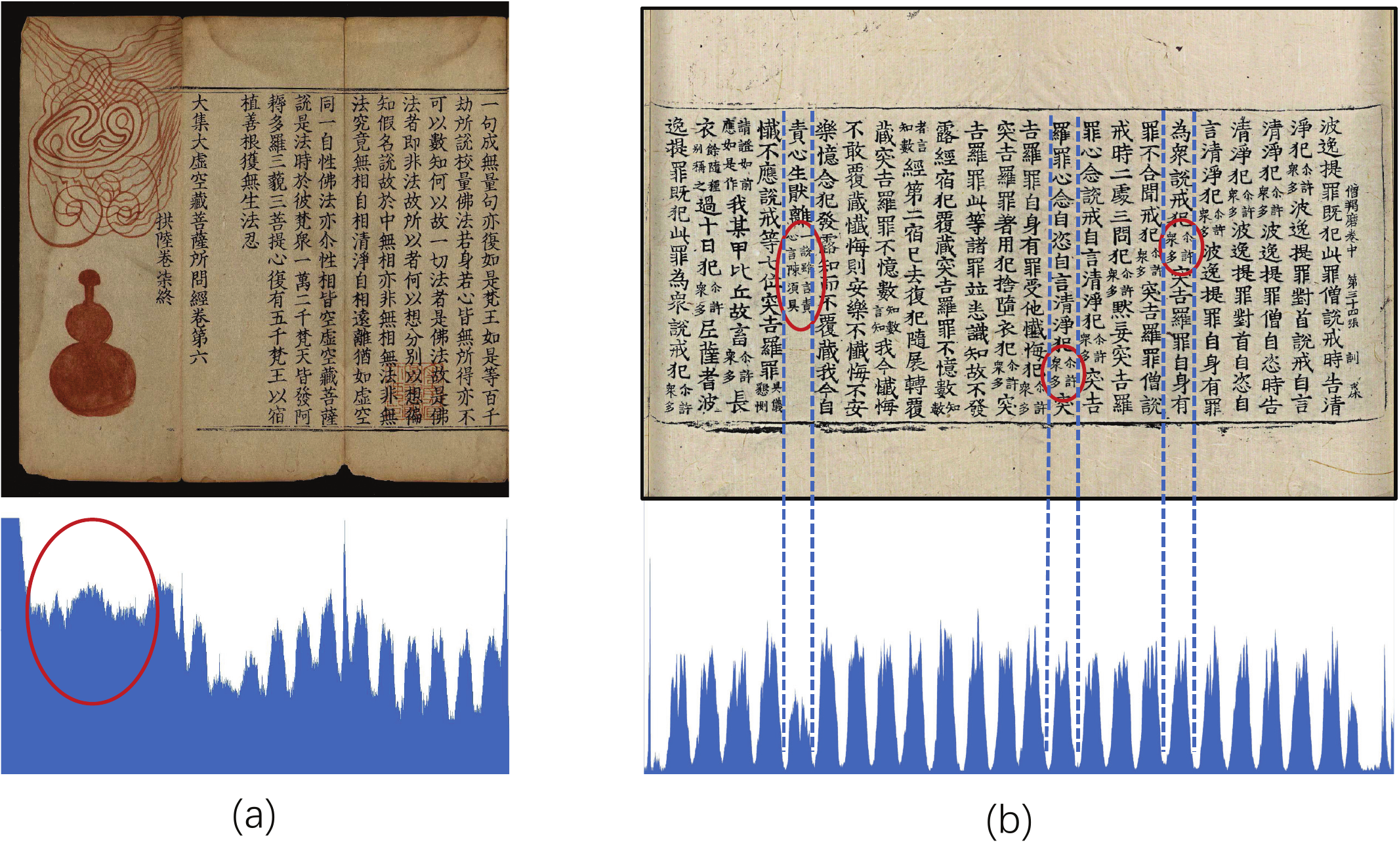}
    \caption{Failures of projection analysis to the segment text line. (a) Complex background. (b) Double-column text sandwiched by single-column text.}
    \label{fig:failure_use_projection}
\end{figure}

Traditional methods mainly rely on hand-crafted features\cite{li2007skew,nagy1992prototype,breuel2002two} and are not robust on historical documents.
As shown in Fig. \ref{fig:failure_use_projection}a, projection profile analysis\cite{li2007skew} fails for images with complex backgrounds.
Moreover, it may lead to an incorrect result when a double-column text is sandwiched by a single-column text, as shown in Fig. \ref{fig:failure_use_projection}b.
With the development of deep learning in recent years, significant progress has been made in these two stages (layout analysis and text recognition)\cite{chen2017convolutional, wick2018fully, tang2016cnn, nguyen2019character}.
However, these processes are still handled separately in a document digitization system.
The overall pipeline is also complicated and time-consuming.
Moreover, errors will accumulate in the process of document digitization if the performance of layout analysis is poor.

\begin{figure*}[htb]
    \centering
    \includegraphics[width=12cm]{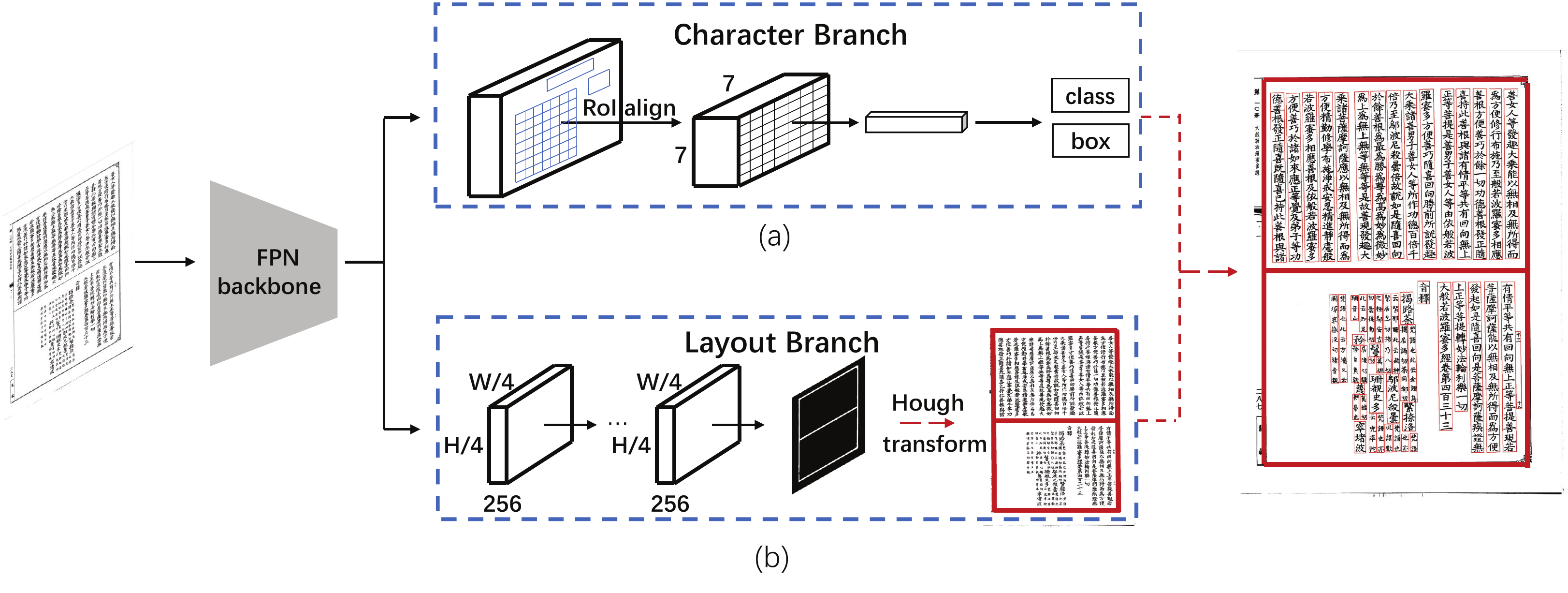}
    \caption{The overall architecture of the proposed framework: 
    The Feature Pyramid Network (FPN) architecture operates on features extracted by ResNet-50.
    (a) Character branch is used for character classification and bounding box regression.
    (b) Layout branch uses a fully convolutional network to generate a binary mask. Hough transform is then adopted to detect lines on the output binary mask.
    Using the results obtained from these two branches, we adopt post-processing method to group characters belonging to the same column and generate text line recognition results.
    Finally, the document is output following the correct reading order.
    The red dashed arrows indicate the steps only in the inference period.}
    \label{fig:overall-architecture}
\end{figure*}

In this paper, we propose a novel framework for conducting layout analysis, character detection, and recognition simultaneously.
Inspired by Mask R-CNN\cite{he2017mask}, which extends Faster R-CNN\cite{ren2015faster} by adding a branch for predicting the segmentation mask, we also add a branch for layout analysis.
As shown in Fig. \ref{fig:overall-architecture}, the layout branch adopts a fully convolutional network (FCN) to output a binary mask, which is taken for line detection in the process of layout analysis.
Experiments show the FCN method is more robust than hand-crafted features.
The recognition branch is character-level.
Thus we can group characters in the vertical direction and avoid the situation where the extracted text line contains the double-column text.
Combining layout information and character results, 
post-processing method is adopted to group characters belonging to the same column and generate text line recognition results.
Finally, we output the document following the reading order.
Furthermore, we propose a re-score mechanism to reduce the impact of document degradation on recognition by taking advantage of the implicit language model.

To summarize, the contributions of this work are three-fold:
\begin{itemize}
\item We propose an end-to-end trainable framework that performs layout analysis, character detection and recognition simultaneously.
\item We propose a re-score mechanism to take advantage of the implicit language model to predict damaged characters, which further boosts recognition performance.
\item To facilitate the research in Chinese historical documents, we extend the size of the original Chinese historical dataset\cite{yang2018dense} with layout, characters, and text lines annotation. The dataset is now available$\footnote[1]{https://github.com/HCIILAB/MTHv2\_Datasets\_Release}$.
\end{itemize}

\section{Related work}

In this section, we review previous works on layout analysis and text recognition.
For document text recognition, we focus on character-based methods.

Classical layout analysis methods can be divided into two categories: top-down and bottom-up approaches.
Top-down approaches such as projection profile analysis\cite{li2007skew}, recursive x-y cuts\cite{nagy1992prototype} and white space analysis\cite{breuel2002two}, start from the whole image and then find components recursively.
This process will not stop until the component finds homogeneous regions.
Bottom-up approaches such as run-length smearing algorithm\cite{wong1982document}, and Voronoi diagram-based algorithm\cite{kise1998segmentation}, take the pixels or connected components as the basic element, then merge the components with similar features into larger homogeneous regions.
Although these methods are simple and efficient for some specific situations, most of them cannot be easily generalized to various layout situations.
Recently, deep-learning-based methods have shown great power in the field of page segmentation.
Chen et al.\cite{chen2017convolutional} proposed a simple architecture that contains only one convolution layer for page segmentation.
The image patches were first generated by using the superpixels algorithm and then were classified into different classes.
Their results outperformed traditional methods.
Wick et al.\cite{wick2018fully} classified the image pixels into the background and other text types using a U-Net\cite{ronneberger2015u} structure model.
This method does not require additional pre-processing steps such as superpixel generation.
In these cases, the layout analysis problem is considered as a pixel-level classification problem.

Historical document character recognition has been a challenging topic due to low image quality, variety of writing styles and lack of labeled training samples.
Li et al.\cite{li2014historical} focused on the problem of various styles and used style transfer mapping method, which allowed the classifier trained by available printed Chinese characters to work on historical recognition.
The proposed method improves the recognition accuracy when characters are of various styles.
Aiming to address the problem of lacking sufficient labeled training samples, Tang et al.\cite{tang2016cnn} adopted the transfer learning method.
A CNN model was first trained using printed Chinese characters samples in the source domain, and the weight was used to initialize another CNN model which was then fine-tuned by a few labeled historical characters.
For the problem of low image quality, Nguyen et al.\cite{nguyen2019character} proposed a character attention generative adversarial network to create high visibility images from severely degraded or low visibility input images.
The recognition rate showed that recognition from the generated images was much better than that from degraded images.

To the best of our knowledge, the method described in this work is the first end-to-end framework for jointly outputting layout analysis, character detection and recognition results.

\section{METHODOLOGY}

\subsection {Overall Architecture}
The proposed framework, which is end-to-end trainable, can output the document layout information, character detection and classification results.
Combining these three types of results, we output the document following the correct reading order.
The overall architecture of our method is presented in Fig. \ref{fig:overall-architecture}.
Functionally, the framework consists of three components: ResNet-50\cite{he2016deep} with feature pyramid network (FPN)\cite{lin2017feature} as backbone, layout branch for generating binary mask containing a region of lines, character branch for generating character detection and classification results.

\subsection{Backbone network}
Characters in historical documents vary in different sizes. 
To extract more robust features at different scales, we use a ResNet-50 with feature pyramid structure\cite{lin2017feature}.
FPN uses a top-down architecture with lateral connections to fuse features of different resolutions from a single-scale input, which improves accuracy with marginal cost.

\subsection{Layout branch}
Layout information is essential in the process of document digitization.
As shown in Fig. \ref{fig:layout_branch},
different layouts in a document image can be distinguished by line segments which are called the boundary lines later.
It is the characteristic in TKH and MTH datasets\cite{yang2018dense}.
Therefore, we consider distinguishing different layouts by boundary lines in this work.
The proposed layout branch based on FCN can classify pixels into lines or non-line region.

\subsubsection{Network structure}
Assuming the shape of input image is $H_i \times W_i$, the shape of the feature map from backbone network is $H_i/4 \times W_i/4$.
The layout branch has the following network structure:

256C3 - BN - ReLU - dropout - 256C3 - BN - ReLU - dropout - 2C1,
where xCy represents a convolutional layer with kernel size of y$\times$y and output channels of x, BN, dropout means batch normalization layer\cite{ioffe2015batch} and dropout layer\cite{srivastava2014dropout}, respectively.
The output binary mask includes two channels, which indicates lines or non-line pixels.

\subsubsection{Label generation}
\begin{figure}
    \centering
    \includegraphics[width=0.35\textwidth]{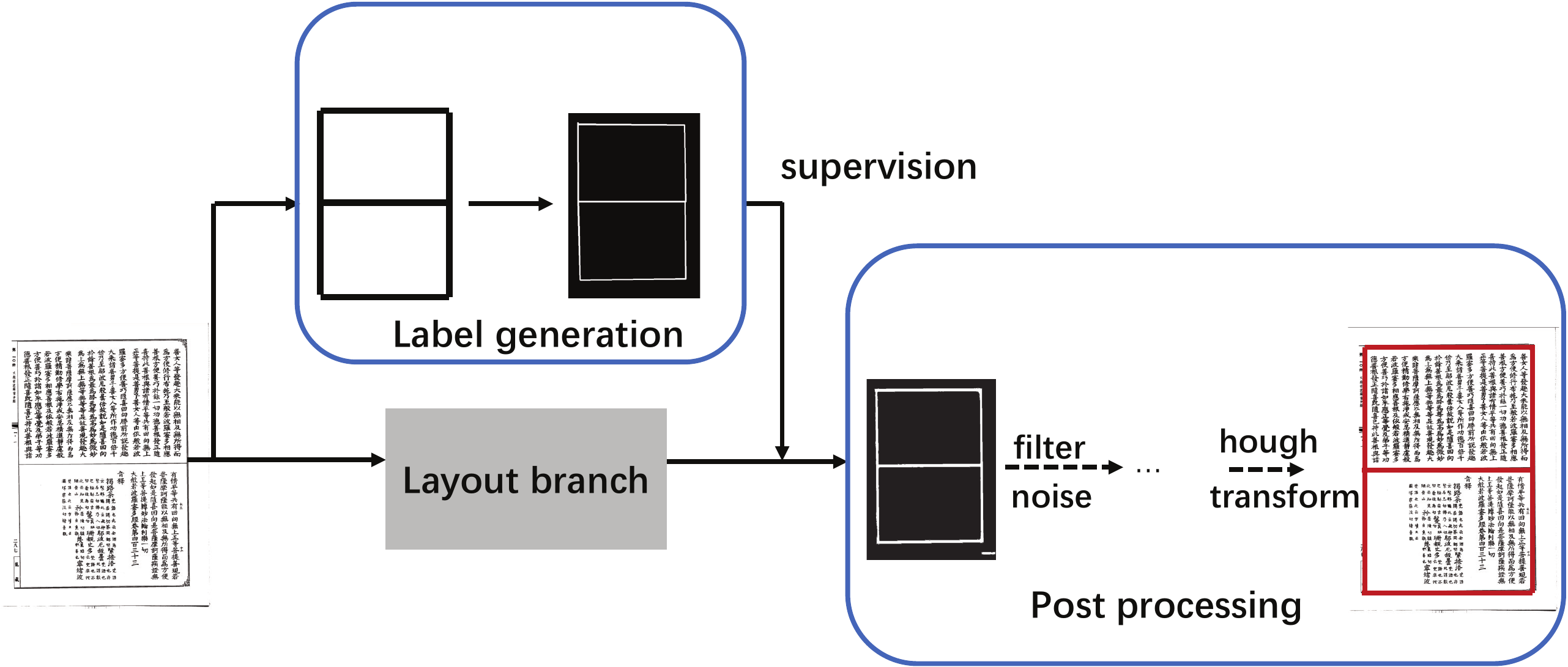}
    \caption{The pipeline of the Layout branch. Pixel-level supervision is generated from boundary lines. In the testing phase, we filter noise and do Hough transform on the binary output. The dashed arrows indicate the steps only in the testing phase.}
    \label{fig:layout_branch}
\end{figure}

Since pixel-level annotation is expensive, pixel-level annotations of the layout are not provided in the MTHv2 datasets.
Therefore, we extend the original dataset annotations. 
As shown in Fig. \ref{fig:layout_branch},
we annotate boundary lines in a document image.
Each boundary lines are represented by their start and end points.
In the training phase, we connect the start and end points of the boundary lines, and the pixels within 20 pixels from the boundary lines will be treated as positive pixels.
Through the above operation, we obtain the binary mask as the label.

\subsubsection{Post-processing}
In the testing phase, the branch outputs binary mask with 1/4 of the original size.
Connected components whose area is smaller than a set threshold are considered as noise output.
We first filter noise output and resize the binary mask back to the original size.
To extract boundary lines, we use Hough line transform\cite{duda1972use} on the binary mask.
There will be a lot of redundant detected lines in this process and we filter them by analyzing their intercept and slope.
The intercept threshold is set to 200 pixels, and the slope threshold is set to \ang{45} in this work.

\subsection{Character branch}
In parallel with the layout branch, the character branch localizes individual characters and recognize them simultaneously.
Compared with methods regressing text line, the character-based method can make sure that the text line grouped by characters only contains the single-column text.
At the same time, the detection and recognition tasks are highly correlated
and complementary.
These two tasks can achieve a better performance if they are trained in a unified network.
It has been proved in the field of scene text\cite{lyu2018mask, Xing_2019_ICCV}.

Following the design of Faster R-CNN, we have two stages in the character branch.
The first stage is a Region Proposal Network (RPN) which generates candidate object proposals.
The second stage is Fast R-CNN\cite{girshick2015fast}, which extracts features using RoIPool from each candidate proposal and performs classification and bounding-box regression.
In our implementation, we adopt RoIAlign\cite{he2017mask} rather than RoIPool.
The input of our framework is a full-page image, and the output format is $(x_{left}, y_{top}, x_{right}, y_{bottom}, cls)$, which means left, top, right, bottom coordinate of the detected character bounding box and classification results, respectively.
Based on character results, we adopt a simple post-processing method to group characters belonging to the same column and generate text line results.

The post-processing method is described as follows.
We first use layout information to find characters that fall in the same layout.
We then group characters into the same column in the vertical direction if their left or right coordinates are smaller than the threshold.
The columns obtained above are still insufficiently accurate when a double-column line of text is sandwiched by a single-column line of text, as shown in Fig. \ref{fig:failure_use_projection} (b).
Therefore, for each column obtained above, we pick up small characters.
And we continue to group these small characters if they belong to the same column in the vertical direction.
Finally, we output the recognition results in these columns following the reading order.

\subsection{End-to-End training}
As discussed above, our framework includes multiple tasks.
A multi-task loss function for end-to-end training is defined:
\begin{equation}
    L=L_{cls}+L_{bbox}+\lambda L_{layout}
\end{equation}
where $L_{cls}$ and $L_{bbox}$ are characters classification loss and bounding box regression loss, respectively, which are identical as those defined in \cite{ren2015faster}.
In addition, $L_{layout}$ is used to optimize layout branch and $\lambda$ is set to 1.0 in the experiments.
\begin{equation}
    L_{layout} = \frac{1} {N} \sum_{i=1}^{N} -log(\frac{e^{p_{i}}}{\sum_{j}{e^{p_{j}}}})
\end{equation}
where $p_{i}$ is confidence of target class, $j$ is the class indice and $N$ is the batch size.
The loss is minimized using stochastic gradient descent (SGD).

\subsection{Re-score mechanism}
In this section, we introduce the re-score mechanism to re-assign the output confidence from the character branch.
As claimed by Xie et al.\cite{xie2017learning}, a text line recognition network can induce the implicit language model for text recognition by using residual LSTM\cite{xie2017learning} to learn long-term context information.
Inspired by this, we propose a re-score mechanism to reduce the impact of degradation on document recognition.

Concretely, a CRNN-based\cite{shi2016end} text line model is trained using text line annotated data.
After character network inference, 
we group characters into columns and get sequence output $R_{c}$ = \{$c_{1}$,$c_{2}$,...,$c_{n}$\} and $P_{c}$ = \{$p_{1}$, $p_{2}$,...,$p_{n}$\}, where $R_{c}$ represents the recognition results from characters belonging to the same column, and $P_{c}$ represents their output probabilities.
Using the column obtained above as the input of the text line model, we also get the sequence text line recognition results $R_{t}$ and sequence probabilities $P_{t}$, respectively.

\begin{figure}[h]
    \centering
    \includegraphics[width=0.48\textwidth]{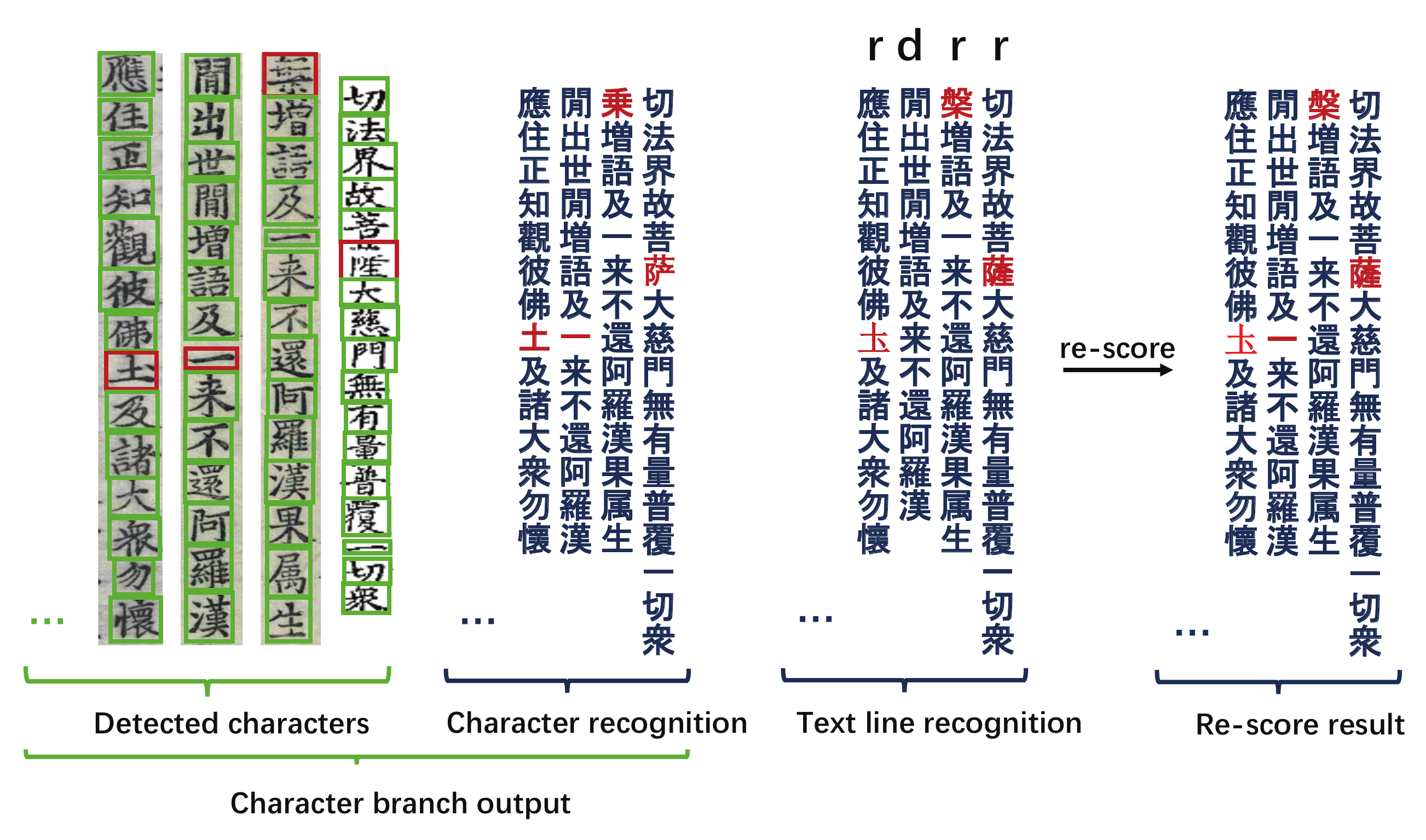}
    \caption{Example output when using re-score mechanism. Text line recognition result is compared with character recognition result based on edit distance. Here r, d represent "\textit{replace}" and "\textit{delete}", respectively.}
    \label{fig:re_score}
\end{figure}

We use the edit distance algorithm to compare strings $R_{t}$ and $R_{c}$.
All operations in the edit distance algorithm include \textit{insert}, \textit{delete}, \textit{replace} or \textit{equal}.
If the operation is only ``\textit{replace}'' when the result of text line is compared with the character recognition result, we then iterate over the mismatched characters and determine the final output based on each probability.
If the operation required includes more than ``\textit{replace}'', we determine the final output based on their average probabilities.
In other cases, the final output is the same as the character recognition result.
As shown in Fig. \ref{fig:re_score},
this proposed mechanism can re-assign the correct output when we predict damaged characters.
Meanwhile, as shown in the second column in Fig. \ref{fig:re_score},
the operation is ``\textit{delete}'' when text line recognition result is compared with character result.
As described above, the final output is the same as the character recognition output.
It helps to avoid situations where text line model sometimes predicts fewer or more characters.

\section{Experiments}

\subsection{Dataset}
The dataset used in this work is the Tripitaka Koreana in Han (TKH) Dataset and the Multiple Tripitaka in Han (MTH) Dataset, which were introduced in paper\cite{yang2018dense}.
To facilitate the research in Chinese historical documents, more challenging document images from the Internet are added to MTH dataset whose images number is now 2200, the combined dataset of TKH and MTH2200 is named MTHv2.
Details of the dataset are given in Table \ref{table:MTHv2 dataset}.

In this dataset, we provide three types of annotations.
The first type is line-level annotation, including text line location and its transcription, which is saved in reading order.
The second type is character-level annotation, which includes class categories and bounding box coordinates.
The last type is the boundary lines, represented by the start and end points of line segments.
We randomly split the MTHv2 dataset into training set and testing set with the ratio of 3:1.
The specific training set and test set are also be published.

\begin{table}[h]
    \footnotesize
    \centering
    \caption{Details of the MTHv2 datasets}
    \label{table:MTHv2 dataset}

    \begin{tabular}[!hbt]{|c|c|c|}
        \hline
         & TKH dataset          & MTH2200 dataset \\
        \hline
        page numbers           &  999     &  2200  \\
        \hline
        text line numbers       &  23472    &  82107  \\
        \hline
        character numbers      &  323502   &  758161  \\
        \hline
        character categories   &  1492     &  6711  \\
        \hline

    \end{tabular}
\end{table}

\subsection{Implementation details}
In the character branch implementation, following FPN\cite{lin2017feature}, we assign anchors on different stages depending on their size.
Specifically, the area of the anchors are set to \{$32^{2}$, $64^{2}$, $128^{2}$, $256^{2}$, $512^{2}$\} pixels on five stages \{$P_{2}$, $P_{3}$, $P_{4}$, $P_{5}$, $P_{6}$\}, respectively.
We also use anchors of multiple aspect ratios \{1:2, 1:1, 2:1\} at each level as in \cite{ren2015faster}.
We adopt RoIAlign to sample the proposals to a 7x7 size.
The number of character classes is 6762, which is set in the last fully connected layer.

In general, document images are at a very large scale and characters are densely distributed in Chinese historical document images.
Therefore, in the training phase, we first scale images to a fixed size, and we use sliding windows methods to crop the image with the overlap size of $100 \times 100$.
In the testing phase, we slide on the images and concatenate these outputs.
Finally, NMS is utilized to suppress redundant character boxes.
With the backbone pre-trained on ImageNet, the network is optimized with stochastic gradient descent(SGD) and the maximum iteration is 40k with the batch size of 1.

Our text line recognition network has the following network architecture:

16C3 - MP2 - 64C3 -MP2 - 128C3 - 256C3 - MP2 - 256C3 - MP2 - 512C3*2 - MP2 - BN - ResidualLSTM*3 - FC6762 - CTC,
where xCy represents a convolutional layer with kernel size of y $\times$ y and output channels of x, MPx denotes a maximum pooling layer with kernel size of x, FCx means a fully connected layer with output number of x.
To get the sequence probabilities, we add a softmax layer to the last layer.
The height of input text line is 96 and the width is kept proportional to the aspect ratio.
The model is optimized with SGD and trained with batch size of 32.

\subsection{Layout analysis evaluation}
In this section, 
the performance of the layout branch is evaluated in terms of Precision, Recall, and F-score.
Since the type of output is a line segment, we define the criteria to determine positive and negative output.
Each detected line is matched to a ground-truth line that has the minimum distance between the start and end points.
Only detected line whose distance value is smaller than a set threshold is thought as a positive output and the others are recorded as negative.

\begin{figure}[b]
    \centering
    \includegraphics[width=0.40\textwidth]{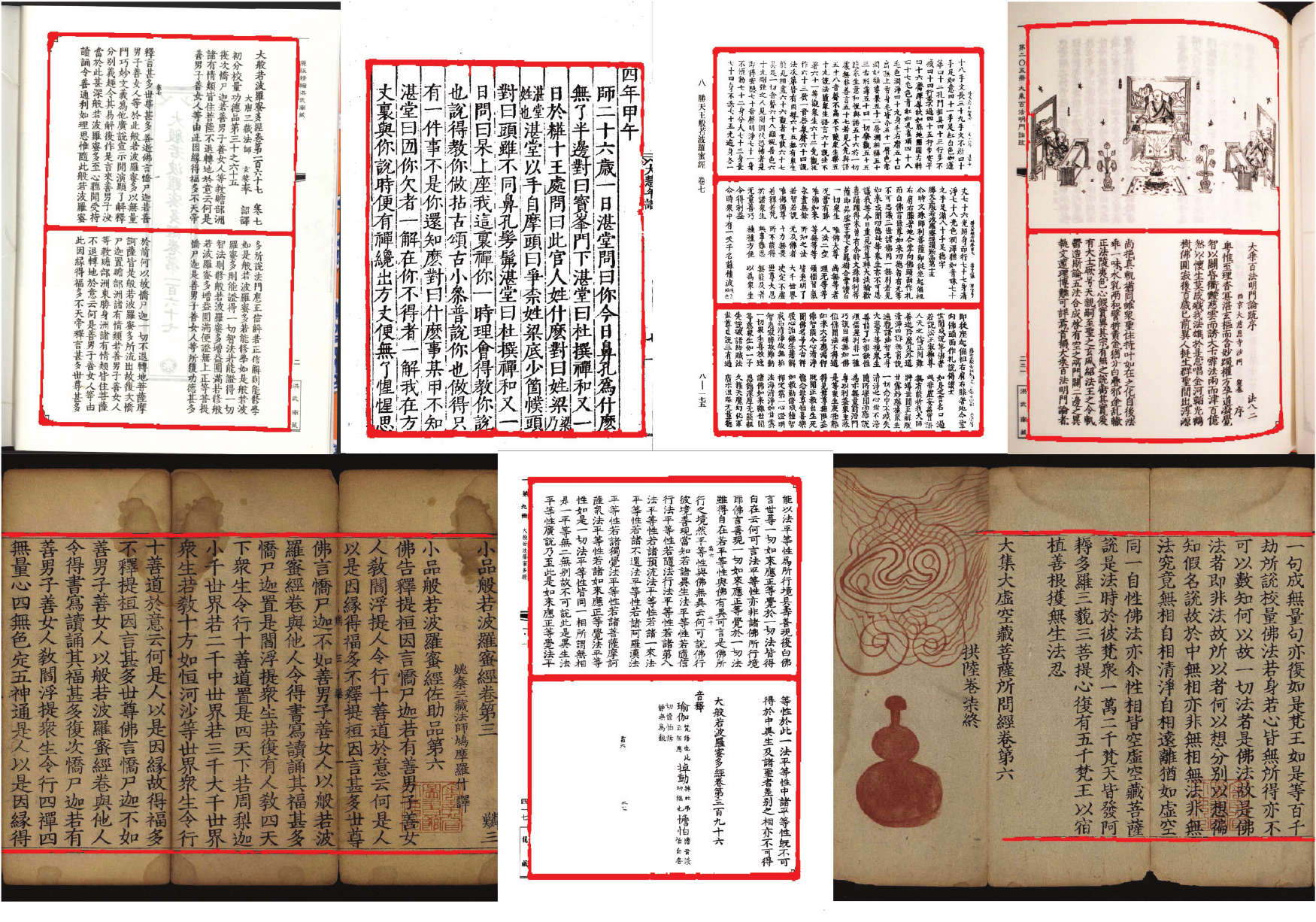}
    \caption{Example output from the layout branch. Colors in red means the lines prediction region.}
    \label{fig:demo_pred}
\end{figure}

As claimed in \cite{yang2018dense}, characters in the TKH dataset are neatly arranged and can be directly output without considering layout information.
To show the superiority of our method, we filter the TKH images in the test set and evaluate complex layout documents, MTH2200.
For comparison, we compare our method with projection analysis.
In our implementation, we first do image pre-processing to erode the noise of background and then do projection analysis.
The pre-processing is that use the horizontal kernel and vertical kernel to erode on the image, which can preserve horizontal lines and vertical lines, respectively.
As shown in Table \ref{table:layout analysis},
our method is significantly better than the traditional method, which has a 20.43\% improvement with F-score.
The high precision shows that the FCN method is more robust on various documents with complex backgrounds and layouts.
Some example output is shown in Fig. \ref{fig:demo_pred}.

\begin{table}[h]
    \footnotesize
    \centering
    \caption{Effectiveness of layout branch on MTH2200 dataset}
    \label{table:layout analysis}
    \begin{tabular}{|c|c|c|c|}
        \hline
        Method & Precision(\%) & Recall(\%) & F-score(\%)\\
        \hline
        Projection analysis & 71.57 & 86.12 & 78.17 \\
        \hline
        Our method  & \textbf{97.77} & 99.44 & \textbf{98.60} \\
        \hline
    \end{tabular}
\end{table}

\subsection{Text line detection evaluation}
The H-mean under different IoU is calculated to evaluate the text line detection performance.
In Table \ref{table:text line det}, we compare our approach with some other well-established methods, including traditional methods such as projection analysis and deep-learning-based methods.
Since the annotation format of text line is a quadrangle, we use the framework that supports the prediction of quadrangle as a comparative experiment, such as  EAST\cite{zhou2017east} and Mask R-CNN\cite{he2017mask}, which has proven to be successful in other fields.

\begin{figure}[t]
    \centering
    \includegraphics[width=0.35\textwidth]{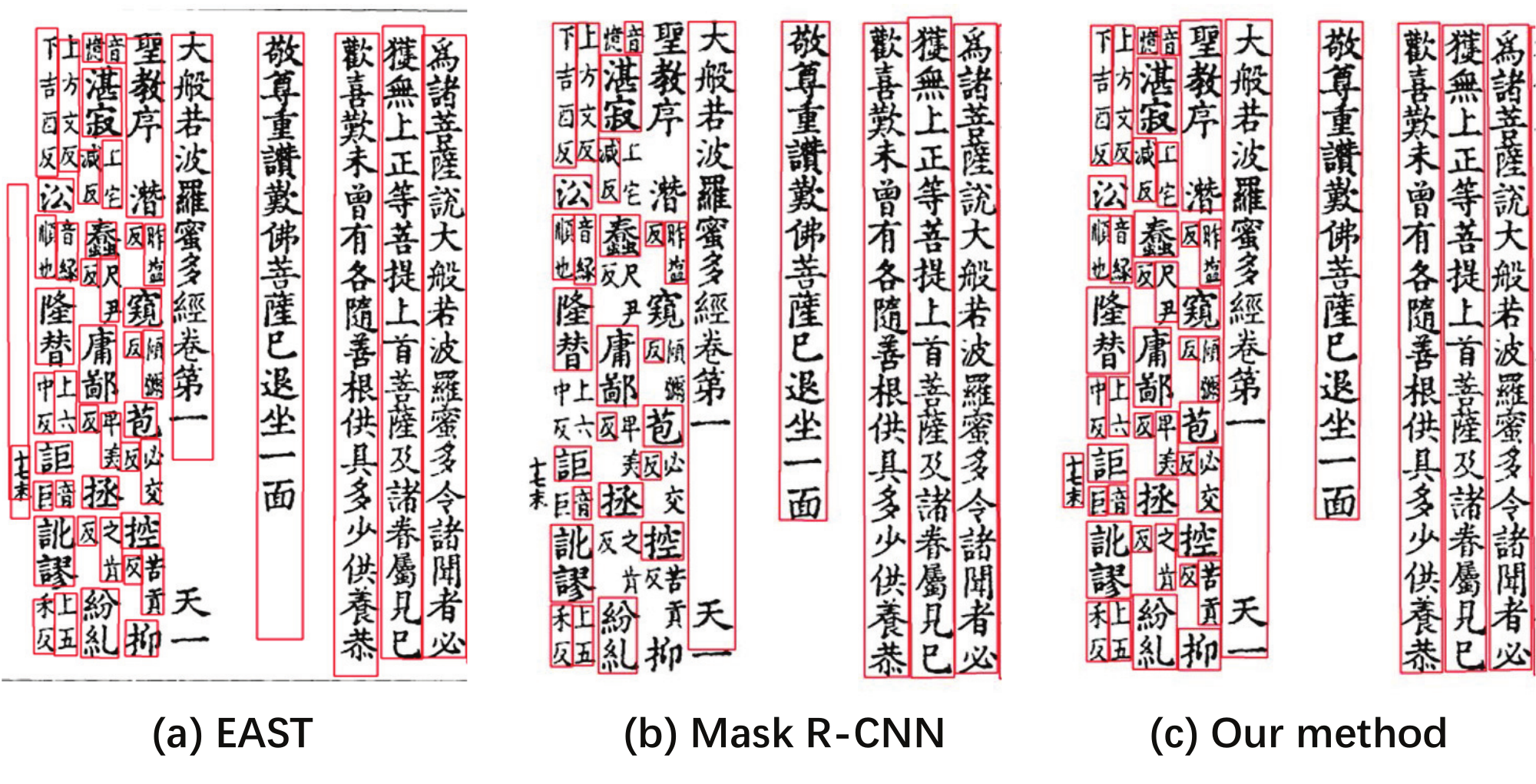}
    \caption{Visualization of detection result, which are the output from EAST, Mask R-CNN and our method, respectively. In our method, we group characters into text line based on character results.}
    \label{fig:demo_det}
\end{figure}

We do not adopt the sliding window method when regressing text lines, because the process of sliding window is easy to cut long text lines into two parts.
Different from our method, these methods which directly regress text line only classify the text line into two categories, text or non-text line, while our method needs to do the classification task of 6762 classes.
As demonstrated in Table \ref{table:text line det}, traditional methods obtain the poor result even when the IoU threshold is 0.5.
This is due to the complex background in the MTHv2 dataset.
The performance of EAST decreases severely with the increase of IoU, which shows it fails to get a tighter bounding box.
Methods such as Mask R-CNN and our character-based method, perform well on document text, and our method reaches the state-of-the-art performance when the IoU threshold range from 0.5 to 0.7.
Besides, the detection and recognition tasks in a unified network can help reject false positives, as claimed in \cite{lyu2018mask}.
Example output is shown in Fig. \ref{fig:demo_det},
which shows our character-based method can get more accurate and tighter bounding boxes.

\begin{table}[htbp]
    \footnotesize
    \centering
    \caption{H-mean result for text line detection}
    \label{table:text line det}
    \begin{tabular}{|c|ccc|}
        \hline
        Method &IoU=0.5(\%)&IoU=0.6(\%)&IoU=0.7(\%) \\
        \hline
        Projection analysis & 69.22       &  66.87     &   60.97  \\
        \hline
        EAST\cite{zhou2017east}  &  95.04 &  91.55     &   80.35  \\
        \hline
        Mask R-CNN\cite{he2017mask}  &   95.55   &  95.18   & 94.27 \\
        \hline
        Our method  &  \textbf{97.72}   &  \textbf{97.26}   & \textbf{96.03}  \\
        \hline
    \end{tabular}
\end{table}

\subsection{Effectiveness of re-score mechanism}
In this section, we use the correct rate (CR) and accuracy rate (AR)\cite{yin2013icdar} to evaluate the performance of our proposed framework.
Concretely, we use both layout analysis and character results to restore documents in the correct reading order (from right to left and top to bottom).
The value of CR and AR are then calculated by comparing the edit distance between the string of ordered output and the string of ground-truth structured document.

The effectiveness of the proposed re-score mechanism on the whole test set is shown in Table \ref{table:re-score}.
The performance of character-based recognition and text line based recognition is comparable.
By using the proposed re-score mechanism, the model reaches the performance with CR of 96.07\% and AR of 95.52\%.
It shows that these two models can complement each other well.
Using the re-score mechanism, we can effectively utilize the implicit language model from the text line model.
Note that the test set in this experiment is different from that in Yang et al.\cite{yang2018dense}, and CR, AR is calculated based on page-level in this work.

\begin{table}
    \footnotesize
    \centering
    \caption{Effectiveness of re-score mechanism}
    \label{table:re-score}
    \begin{tabular}{|c|c|c|c|}
        \hline
        Method & Text line recognition & Character recognition & Re-score \\
        \hline
        CR(\%) & 95.09 & 94.63 & \textbf{96.07} \\
        \hline
        AR(\%) & 94.08 & 94.12 & \textbf{95.52} \\
        \hline
    \end{tabular}
\end{table}

\section {Conclusion}
In this paper,
we propose a novel framework that can simultaneously conduct layout detection, character detection and recognition.
Furthermore, we also propose a mechanism to improve recognition performance with the implicit language model.
Experiments show our framework can outperform traditional and deep-learning-based methods.
Character-level annotations are more expensive than text line annotations.
In the future, we plan to train our framework in a weakly-supervised manner, which has proven to be successful in the scene text field.

\section {Acknowledgement}
This research is supported in part by NSFC (Grant No.: 61936003), GD-NSF (no.2017A030312006), the National Key Research and Development Program of China (No. 2016YFB1001405), Guangdong Intellectual Property Office Project (2018-10-1), and Fundamental Research Funds for the Central Universities (x2dxD2190570).

\bibliographystyle{ieeetr}
\bibliography{ref_v5}

\begin{thebibliography}{10}

\bibitem{ding2004document}
X.~Ding, D.~Wen, L.~Peng, and C.~Liu, ``Document digitization technology and
  its application for digital library in china,'' in {\em First International
  Workshop on Document Image Analysis for Libraries, 2004. Proceedings.},
  pp.~46--53, IEEE, 2004.

\bibitem{xie2019weakly}
Z.~Xie, Y.~Huang, L.~Jin, Y.~Liu, Y.~Zhu, L.~Gao, and X.~Zhang, ``Weakly
  supervised precise segmentation for historical document images,'' {\em
  Neurocomputing}, vol.~350, pp.~271--281, 2019.

\bibitem{li2007skew}
S.~Li, Q.~Shen, and J.~Sun, ``Skew detection using wavelet decomposition and
  projection profile analysis,'' {\em Pattern recognition letters}, vol.~28,
  no.~5, pp.~555--562, 2007.

\bibitem{nagy1992prototype}
G.~Nagy, S.~Seth, and M.~Viswanathan, ``A prototype document image analysis
  system for technical journals,'' {\em Computer}, vol.~25, no.~7, pp.~10--22,
  1992.

\bibitem{breuel2002two}
T.~M. Breuel, ``Two geometric algorithms for layout analysis,'' in {\em
  International workshop on document analysis systems}, pp.~188--199, Springer,
  2002.

\bibitem{chen2017convolutional}
K.~Chen, M.~Seuret, J.~Hennebert, and R.~Ingold, ``Convolutional neural
  networks for page segmentation of historical document images,'' in {\em 2017
  14th IAPR International Conference on Document Analysis and Recognition
  (ICDAR)}, vol.~1, pp.~965--970, IEEE, 2017.

\bibitem{wick2018fully}
C.~Wick and F.~Puppe, ``Fully convolutional neural networks for page
  segmentation of historical document images,'' in {\em 2018 13th IAPR
  International Workshop on Document Analysis Systems (DAS)}, pp.~287--292,
  IEEE, 2018.

\bibitem{tang2016cnn}
Y.~Tang, L.~Peng, Q.~Xu, Y.~Wang, and A.~Furuhata, ``Cnn based transfer
  learning for historical chinese character recognition,'' in {\em 2016 12th
  IAPR Workshop on Document Analysis Systems (DAS)}, pp.~25--29, IEEE, 2016.

\bibitem{nguyen2019character}
K.~C. Nguyen, C.~T. Nguyen, S.~Hotta, and M.~Nakagawa, ``A character attention
  generative adversarial network for degraded historical document
  restoration,'' in {\em 2019 International Conference on Document Analysis and
  Recognition (ICDAR)}, pp.~420--425, IEEE, 2019.

\bibitem{he2017mask}
K.~He, G.~Gkioxari, P.~Doll{\'a}r, and R.~Girshick, ``Mask r-cnn,'' in {\em
  Proceedings of the IEEE international conference on computer vision},
  pp.~2961--2969, 2017.

\bibitem{ren2015faster}
S.~Ren, K.~He, R.~Girshick, and J.~Sun, ``Faster r-cnn: Towards real-time
  object detection with region proposal networks,'' in {\em Advances in neural
  information processing systems}, pp.~91--99, 2015.

\bibitem{yang2018dense}
H.~Yang, L.~Jin, W.~Huang, Z.~Yang, S.~Lai, and J.~Sun, ``Dense and tight
  detection of chinese characters in historical documents: Datasets and a
  recognition guided detector,'' {\em IEEE Access}, vol.~6, pp.~30174--30183,
  2018.

\bibitem{wong1982document}
K.~Y. Wong, R.~G. Casey, and F.~M. Wahl, ``Document analysis system,'' {\em IBM
  journal of research and development}, vol.~26, no.~6, pp.~647--656, 1982.

\bibitem{kise1998segmentation}
K.~Kise, A.~Sato, and M.~Iwata, ``Segmentation of page images using the area
  voronoi diagram,'' {\em Computer Vision and Image Understanding}, vol.~70,
  no.~3, pp.~370--382, 1998.

\bibitem{ronneberger2015u}
O.~Ronneberger, P.~Fischer, and T.~Brox, ``U-net: Convolutional networks for
  biomedical image segmentation,'' in {\em International Conference on Medical
  image computing and computer-assisted intervention}, pp.~234--241, Springer,
  2015.

\bibitem{li2014historical}
B.~Li, L.~Peng, and J.~Ji, ``Historical chinese character recognition method
  based on style transfer mapping,'' in {\em 2014 11th IAPR International
  Workshop on Document Analysis Systems}, pp.~96--100, IEEE, 2014.

\bibitem{he2016deep}
K.~He, X.~Zhang, S.~Ren, and J.~Sun, ``Deep residual learning for image
  recognition,'' in {\em Proceedings of the IEEE conference on computer vision
  and pattern recognition}, pp.~770--778, 2016.

\bibitem{lin2017feature}
T.-Y. Lin, P.~Doll{\'a}r, R.~Girshick, K.~He, B.~Hariharan, and S.~Belongie,
  ``Feature pyramid networks for object detection,'' in {\em Proceedings of the
  IEEE conference on computer vision and pattern recognition}, pp.~2117--2125,
  2017.

\bibitem{ioffe2015batch}
S.~Ioffe and C.~Szegedy, ``Batch normalization: Accelerating deep network
  training by reducing internal covariate shift,'' {\em arXiv preprint
  arXiv:1502.03167}, 2015.

\bibitem{srivastava2014dropout}
N.~Srivastava, G.~Hinton, A.~Krizhevsky, I.~Sutskever, and R.~Salakhutdinov,
  ``Dropout: a simple way to prevent neural networks from overfitting,'' {\em
  The journal of machine learning research}, vol.~15, no.~1, pp.~1929--1958,
  2014.

\bibitem{duda1972use}
R.~O. Duda and P.~E. Hart, ``Use of the hough transformation to detect lines
  and curves in pictures,'' {\em Communications of the ACM}, vol.~15, no.~1,
  pp.~11--15, 1972.

\bibitem{lyu2018mask}
P.~Lyu, M.~Liao, C.~Yao, W.~Wu, and X.~Bai, ``Mask textspotter: An end-to-end
  trainable neural network for spotting text with arbitrary shapes,'' in {\em
  Proceedings of the European Conference on Computer Vision (ECCV)},
  pp.~67--83, 2018.

\bibitem{Xing_2019_ICCV}
L.~Xing, Z.~Tian, W.~Huang, and M.~R. Scott, ``Convolutional character
  networks,'' in {\em The IEEE International Conference on Computer Vision
  (ICCV)}, October 2019.

\bibitem{girshick2015fast}
R.~Girshick, ``Fast r-cnn,'' in {\em Proceedings of the IEEE international
  conference on computer vision}, pp.~1440--1448, 2015.

\bibitem{xie2017learning}
Z.~Xie, Z.~Sun, L.~Jin, H.~Ni, and T.~Lyons, ``Learning spatial-semantic
  context with fully convolutional recurrent network for online handwritten
  chinese text recognition,'' {\em IEEE transactions on pattern analysis and
  machine intelligence}, vol.~40, no.~8, pp.~1903--1917, 2017.

\bibitem{shi2016end}
B.~Shi, X.~Bai, and C.~Yao, ``An end-to-end trainable neural network for
  image-based sequence recognition and its application to scene text
  recognition,'' {\em IEEE transactions on pattern analysis and machine
  intelligence}, vol.~39, no.~11, pp.~2298--2304, 2016.

\bibitem{zhou2017east}
X.~Zhou, C.~Yao, H.~Wen, Y.~Wang, S.~Zhou, W.~He, and J.~Liang, ``East: an
  efficient and accurate scene text detector,'' in {\em Proceedings of the IEEE
  conference on Computer Vision and Pattern Recognition}, pp.~5551--5560, 2017.

\bibitem{yin2013icdar}
F.~Yin, Q.-F. Wang, X.-Y. Zhang, and C.-L. Liu, ``Icdar 2013 chinese
  handwriting recognition competition,'' in {\em 2013 12th International
  Conference on Document Analysis and Recognition}, pp.~1464--1470, IEEE, 2013.

\end{thebibliography}

\end{document}